\documentclass[runningheads]{llncs}

\usepackage{eccv}

\usepackage{eccvabbrv}
\usepackage{graphicx}
\usepackage{booktabs}

\usepackage[accsupp]{axessibility}  
\usepackage{wrapfig, booktabs, makecell, multirow}
\usepackage{array}

\usepackage[pagebackref,breaklinks,colorlinks,citecolor=eccvblue]{hyperref}

\usepackage{orcidlink}

\begin{document}

\title{GenHOI: Generalized Hand-Object Pose Estimation with Occlusion Awareness} 

\titlerunning{GenHOI}

\author{Hui Yang\inst{1,2}\orcidlink{0009-0005-2648-9566} \and
Wei Sun\inst{1*}\orcidlink{0000-0002-2589-2100} \and
Jian Liu\inst{1*}\orcidlink{0000-0003-0604-8024} \and
Jian Xiao\inst{1}\orcidlink{0009-0002-6919-6550} \and
Tao Xie\inst{1}\orcidlink{0009-0006-3190-1196} \and
\\Hossein Rahmani\inst{3}\orcidlink{0000-0003-1920-0371} \and
Ajmal Mian\inst{4}\orcidlink{0000-0002-5206-3842} \and
Nicu Sebe\inst{5}\orcidlink{0000-0002-6597-7428} \and
Gim Hee Lee\inst{2}\orcidlink{0000-0002-1583-0475}
{\tt\small \{huiyang, wei\_sun, jianliu\}@hnu.edu.cn,  
 gimhee.lee@comp.nus.edu.sg}
}

\authorrunning{H.~Yang et al.}

\institute{
\textsuperscript{1 }NERC-RVC, Hunan University,
\textsuperscript{2 }National University of Singapore,
\\
\textsuperscript{3 }Lancaster University,
\textsuperscript{4 }The University of Western Australia,
\\
\textsuperscript{5 }University of Trento
}
\maketitle

\begin{abstract}
{\let\thefootnote\relax\footnotetext{$^{*}$ Corresponding authors.}}
Generalized 3D hand-object pose estimation from a single RGB image remains challenging due to the large variations in object appearances and interaction patterns, especially under heavy occlusion. We propose GenHOI, a framework for generalized hand-object pose estimation with occlusion awareness. GenHOI integrates hierarchical semantic knowledge with hand priors to enhance model generalization under challenging occlusion conditions. Specifically, we introduce a hierarchical semantic prompt that encodes object states, hand configurations, and interaction patterns via textual descriptions. This enables the model to learn abstract high-level representations of hand-object interactions 
{for generalization} to unseen objects and novel interactions while compensating for missing or ambiguous visual cues. To enable robust occlusion reasoning, we adopt a multi-modal masked modeling strategy over RGB images, predicted point clouds, and textual descriptions. Moreover, we leverage hand priors as stable spatial references to extract implicit interaction constraints. 
{This allows} reliable pose inference even under significant variations in object shapes and interaction patterns. Extensive experiments on the challenging DexYCB and HO3Dv2 benchmarks demonstrate that our method achieves state-of-the-art performance in hand-object pose estimation.
\end{abstract}

\vspace{-2em}

\begin{figure*}[t!]
\centering
\includegraphics[width=1\linewidth]{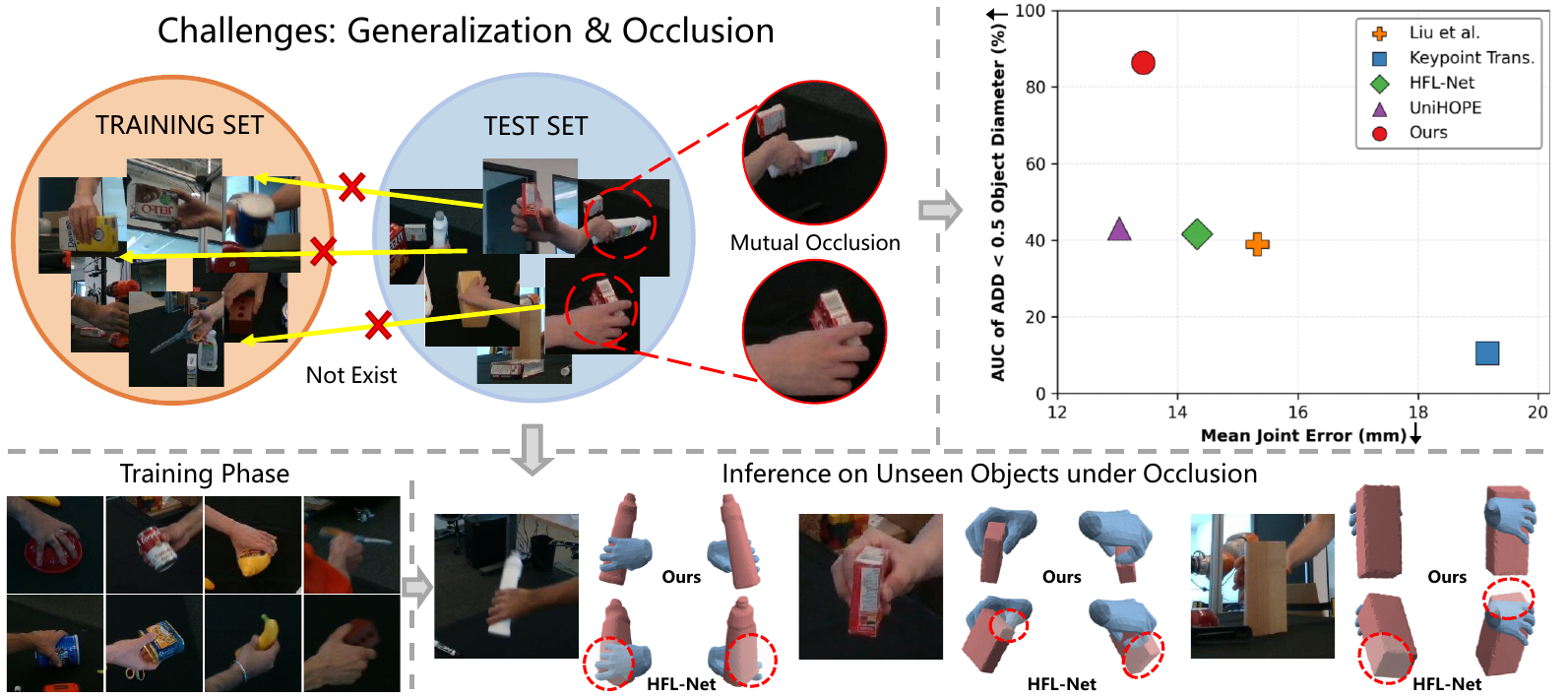}%
\captionof{figure}{Illustration of task challenges and key performance comparisons. Hand-object pose estimation (\emph{left}) faces the challenge of generalizing from training set to unseen test set. Additionally, severe hand-object occlusion makes this task more complicated. 
{Our method shows stronger generalization (\emph{bottom:} qualitative comparison with HFL-Net \cite{lin2023harmonious} on unseen objects in front and back views, and red dotted circles marking focus areas) and achieves the best performance (\emph{right}) on the challenging generalized DexYCB S3 split \cite{chao2021dexycb}. These results confirm effectiveness in generalized settings and under occlusion.}
}
\label{fig1:teaser}
\end{figure*}

\section{Introduction}
\label{sec:intro}

\par Accurate 3D hand-object pose recovery is essential for applications in augmented reality \cite{ar1, ye2023affordance}, human-robot interaction \cite{inter1, inter2, xu2024handbooster}, and robotic manipulation \cite{lj_tii, mani1, diff9d, SinRef6D}. However, real-world hand-object interactions involve a wide variety of objects and interaction scenarios, which hinders the model ability to generalize to unseen objects and interaction patterns. Additionally, frequent occlusions during interactions obscure key visual cues further exacerbating the challenge of generalization in pose estimation.

Most existing hand-object pose estimation methods either directly regress poses from visual features \cite{zhang2024moho, jiang2024hand, wang2025unihope} or estimate them through explicit keypoint correspondences \cite{wang2023interacting, lin2023harmonious, kuang2024learning}. 
{The former relies only on appearance cues to capture surface patterns rather than intrinsic object properties or interaction intent. This limits generalization to unseen categories and diverse interaction scenarios.}
The latter depends on accurate keypoint detection, 
{which makes} it highly sensitive to occlusions and missing visual evidence. In general, {as shown in Fig.~\ref{fig1:teaser},} the bottleneck limiting the application of hand-object pose estimation methods lies in their generalization and robustness to occlusion. 

{To address these limitations, we introduce hierarchical semantic information and interaction priors. Hierarchical semantics provide multi-level descriptions of object properties, hand states, and hand-object interaction patterns. This abstract knowledge supports reasoning about unseen objects and novel interactions, and can supplement missing visual cues. Moreover, the hand provides stable structural priors such as joint positions and rotations during interaction. These internal constraints act as reliable spatial references when object shapes vary widely, which further improves generalization.}
Consequently, fully leveraging semantic cues and interaction priors is essential to overcome the remaining challenges and to improve the generalization of hand-object pose estimation under occlusion.

{In this paper, we propose a framework that integrates hierarchical semantic knowledge with hand priors to improve generalization in hand-object pose estimation, especially under occlusion.}
{Specifically, we introduce a hierarchical semantic prompt that encodes object states, hand configurations, and interaction patterns through textual descriptions. This prompt guides the model toward abstract representations that generalize to unseen objects and novel interactions.}
Additionally, we introduce a multi-modal masked modeling strategy that leverages RGB images, predicted point clouds, and language descriptions. By randomly masking and reconstructing regions within these modalities, the model learns to perceive occlusions and infer missing visual cues. This occlusion-aware learning process enables it to maintain reliable pose estimation even when visual information is partially missing or corrupted. 
{Furthermore, we incorporate hand priors by using joint positions and rotational parameters as stable references that capture implicit spatial constraints during interaction. These priors improve robustness across large variations in object shapes and interaction patterns, and they support reliable pose estimation under severe occlusion.}


\par Our contributions are summarized as follows:
\begin{itemize}
\item We propose a generalized hand-object pose estimation framework tailored to unseen objects and novel interaction patterns under severe occlusion.

\item {We introduce hierarchical semantic prompts that encode object states, hand configurations, and interaction patterns through textual descriptions. This supports pose estimation under occlusion with stronger generalization.}

\item We design a multi-modal masked modeling strategy over RGB images, predicted point clouds, and textual descriptions to learn occlusion-aware representations and infer missing visual cues.

\item {We leverage hand priors as stable spatial references to extract implicit constraints. This helps the model infer object poses across variations in object shapes and interaction patterns, which improves generalization.}
\end{itemize}

\section{Related Work}
\label{sec:Related Work}
\subsection{3D Hand-Object Pose Estimation}

\par Existing hand-object pose estimation approaches can be broadly divided into direct regression and keypoint-based methods. Direct regression methods directly predict the 3D poses of the hand and object from image features. Chen \emph{et al.} \cite{chen2022alignsdf} proposed a unified framework that jointly learns image features and parametric hand-object representations to infer poses. Chen \emph{et al.} \cite{chen2023gsdf} further introduced kinematic constraints to enforce physically plausible hand motion. Qi \emph{et al.} \cite{qi2024hoisdf} designed a fused image-depth representation to jointly regress hand-object configurations, while Zhang \emph{et al.} \cite{zhang2023ddf} incorporated ray-based feature aggregation to better model local hand-object interactions. Keypoint-based approaches estimate the 3D hand-object pose by leveraging spatial relations among detected keypoints. Doosti \emph{et al.} \cite{doosti2020hope} jointly predicted hand and object keypoints and used MANO \cite{mano} and PnP to recover 3D poses. Liu \emph{et al.} \cite{liu2021semi} introduced a semi-supervised framework that utilizes spatiotemporal constraints to provide pseudo-label supervision. Hampali \emph{et al.} \cite{hampali2022keypoint_transformer} integrated cross-attention transformers to explicitly encode joint hand-object geometry, and Lin \emph{et al.} \cite{lin2023harmonious} further enforced dual-stream feature disentanglement for more stable mutual pose refinement.

\par Despite the progress of both paradigms, most existing methods rely primarily on visual cues or fixed keypoint representations and lack semantic and structural priors. As a result, they struggle in heavy-occlusion scenarios and exhibit limited generalization to unseen objects and novel interaction patterns.

\subsection{Textual Semantics for Hand-Object Interaction}

Recent studies \cite{liu2025easyhoi, ye2024ghop} indicate that purely visual cues often struggle under severe occlusion and large intra-class variation, where textual semantic can provide effective complementary guidance. This has motivated the integration of textual semantics into interaction modeling. Zhang \emph{et al.} \cite{zhang2024hoidiffusion} used large language models to produce contextual interaction descriptions and combine them with image generators to synthesize hand-object background images. Following this direction, Cha \emph{et al.} \cite{cha2024text2hoi} introduced a text-guided 3D interaction generation pipeline to produce plausible hand-object configurations. Huang \emph{et al.} \cite{huang2025hoigpt} further build a generative language-to-3D framework that enables interpretable and expressive interaction synthesis. In parallel, textual supervision has been adopted in object pose estimation \cite{liu2024survey, yang2025lancope, yang2025rgb}. Corsetti \emph{et al.} \cite{corsetti2024ov6d} leveraged vision-language priors for open-vocabulary segmentation and 6D pose estimation, while Lin \emph{et al.} \cite{lin2024clipose} leveraged pretrained vision-language models to incorporate semantic knowledge from both image and text modalities for enhanced category-level understanding.
\par {However, these methods mainly rely on textual cues that describe the global scene or coarse object categories, which provide only high-level context.} 
In contrast, our approach tackles the more challenging hand-object pose estimation task, which demands fine-grained spatial reasoning under severe occlusion. We introduce a hierarchical textual representation that encodes the object, the hand, and their interaction 
{to provide} structured multi-level cues for detailed understanding. 
{These hierarchical semantics are integrated with visual and geometric modalities through a multimodal masked modeling strategy, which reconstructs masked signals under random occlusion and improves occlusion awareness and cross-modal reasoning.}

\begin{figure*}[t!]
\centering
\includegraphics[width=1\linewidth]{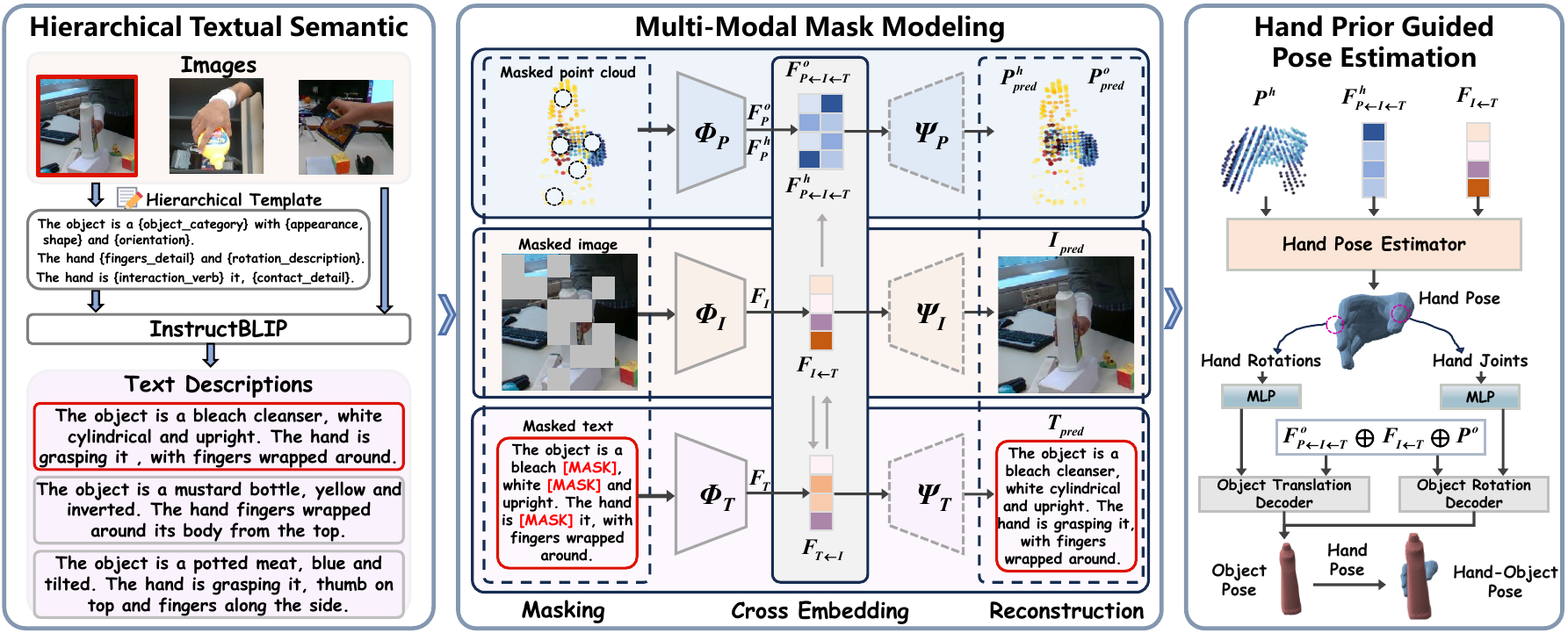}%
\caption{
GenHOI framework, which consists of three key components: 
(1) \textbf{Hierarchical Textual Semantics.} Given RGB images and hierarchical templates with the object, hand, and interaction levels, we employ InstructBLIP to generate hierarchical textual descriptions. 
(2) \textbf{Multi-Modal Mask Modeling.} 
Given the RGB image, textual description, and corresponding hand-object point cloud, we first apply a modality-specific masking strategy to the inputs. The masked inputs are processed by the cross-modal embedding module, which fuses visual, geometric, and textual features to produce representations used for reconstruction and pose estimation. The dashed boxes indicate that masking and reconstruction are used only during training.
(3) \textbf{Hand Prior Guided Pose Estimation.} 
The cross-modal features learned in the previous stage are aggregated to a robust representation. 
{Using these fused features, we first estimate hand pose parameters, which then serve as reliable priors for object pose reasoning.}
}
\label{fig2:framework}
\vspace{-0.25em}
\end{figure*}

\section{Method}

{Fig.~\ref{fig2:framework} shows our proposed GenHOI,}
a novel framework that integrates hierarchical semantic prompts and hand priors to enhance both generalization and robustness under occlusion.

\subsection{Hierarchical Textual Semantic Embedding}
We construct a hierarchical prompt template that encodes multi-level semantics of the hand, object, and their interaction. Leveraging this template with the input image, we generate a unified textual prompt that captures both fine-grained visual details and high-level semantic relationships.

\subsubsection{Hierarchical Prompt Template.}
We define three semantic components to construct the hierarchical prompt template: object-level semantics $s_o$, hand-level semantics $s_h$, and interaction-level semantics $s_i$. 
\par The object-level semantics $s_o$ describe object properties such as category, appearance, and orientation, formulated as  
``the object is a \{object\_category\} with a \{shape, appearance\}, and \{orientation\_description\}.''  
The hand-level semantics $s_h$ characterize hand articulation and grasp configuration, expressed as  
``the hand \{fingers\_detail\} and \{rotation\_description\}.''  
The interaction-level semantics $s_i$ capture the relational context between the hand and the object, following the structure  
``the hand is \{interaction\_verb\} it, \{contact\_detail\}.'' Finally, these three components are integrated into a unified hierarchical prompt $\mathcal T$. This process enables the model to capture a comprehensive multi-level semantic context by linking cues from the object, hand, and their interaction. The final hierarchical template is shown in Fig. \ref{fig2:framework}.

\subsubsection{Hierarchical Textual Semantic Generation.} 
Given an input image $I \in \mathbb{R}^{3 \times H \times W}$ and a hierarchical template $\mathcal T$, 
we leverage a vision-language model 
{such as} InstructBLIP~\cite{dai2023instructblip} 
to generate a textual description $T$. Formally, the generation process is defined as:
\begin{equation}
T = \mathcal{G}(I, \mathcal T),
\end{equation}
where $\mathcal{G}(\cdot)$ denotes the generative vision-language model.
This model fills the template slots with image-specific content 
{to produce} a textual description that aligns the visual observations with the hierarchical semantics encoded in $\mathcal T$. As shown in Fig.~\ref{fig2:framework},
the model generates the hierarchical prompt $\mathcal{T}$ based on the visual content. To improve robustness against potential noise or bias in automatically generated descriptions, textual perturbations are introduced during training to reduce dependence on perfectly consistent inputs. More details are provided in the 
{supplementary material.}
\par This unified hierarchical textual description provides rich semantic information that encapsulates both local and global contextual cues of the hand-object interaction. It is then fused with visual and geometric representations 
{to enable a} more robust and generalizable pose estimation under occlusion.

\subsection{Multi-Modal Masked Modeling}
{Visual cues often become incomplete under severe occlusion due to hand object interaction,} while point cloud geometry and textual semantics can provide complementary information. To enhance occlusion awareness and improve cross-modal consistency, we propose a multi-modal masked modeling strategy that jointly masks and reconstructs visual, geometric, and textual modalities. During training, each image is paired with a point cloud sampled from ground-truth hand and object mesh. 
{During inference, we dynamically sample a point cloud from a voxel grid guided by image features. The sampling follows \cite{qi2024hoisdf, yang2025occlusion} and provides a geometric representation for each image without requiring ground-truth meshes.}
Masking and reconstruction are applied only during training to help the model infer missing information and align multi-modal representations. 
Moreover, the model directly utilizes the learned cross-modal fusion features from image, point cloud, and text, and these integrated representations are then used for subsequent hand-object pose estimation.

\subsubsection{Modality-specific Masking Strategy.} {We apply three modality-specific masks: 1) Image masking corrupts selected patches in the hand-object region and the background. 2) Point cloud masking drops random local patches for both hand and object geometry. 3) Text masking replaces a subset of tokens in the hierarchical description with [MASK].}

\smallskip
\noindent \textbf{Image Masking: } Given an input image $I$, the masked image $\tilde{I}$ is defined as:
\begin{equation}
\begin{aligned}
M_I &= M_{ho}(\lfloor \alpha N_I \rfloor) \cup M_{bg}(N_I - \lfloor \alpha N_I \rfloor), \\
\tilde{I} &= (1 - M_I) \odot I + M_I \odot \mathcal{N}(0, \sigma^2),
\end{aligned}
\end{equation}
where $\odot$ denotes element-wise multiplication, and $M_I \in \{0,1\}^{H \times W}$ is the mask for the image. Here, $M_{ho}$ and $M_{bg}$ are the sets of masked patches sampled from the hand-object region and background region, respectively. $N_I$ is the total number of masked patches and $\alpha \in [0,1]$ controls the proportion of patches masked within the hand-object region. $\mathcal{N}(0, \sigma^2)$ denotes Gaussian noise. Our target-centric masking balances patches from the hand-object region and background, which preserves interaction-focused cues and global contextual information.

\smallskip
\noindent \textbf{Point Cloud Masking: }Given a hand point cloud $P^h \in \mathbb{R}^{3 \times N_h}$, we first partition it into $K$ local patches $\bigcup_{k=1}^{K} \mathbf{P}_k^h = P^h$.
The masked hand point cloud $\tilde{\mathbf{P}}_k^h$ is then defined as:
\begin{equation}
\tilde{\mathbf{P}}_k^h =
\begin{cases}
\mathbf{0}, & k \in \mathcal{K}^h, \\
\mathbf{P}_k^h, & k \notin \mathcal{K}^h,
\end{cases}
\end{equation}
where $\mathcal{K}^h$ is the set of patch indices randomly selected according to the masking ratio $\beta \in [0,1]$. The object point cloud $P^o$ is masked in the same manner as hand, yielding the final masked point clouds $\tilde{P}^h$ and $\tilde{P}^o$.

\smallskip
\noindent \textbf{Textual Masking: } 
Given a hierarchical textual description $\mathbf{T} = \{t_l\}_{l=1}^L$ of length $L$, we randomly select a subset of token positions $\mathcal{L}$ to mask according to a masking ratio $\gamma \in [0,1]$. 
The masked sequence $\tilde{\mathbf{T}} = \{\tilde{t}_l\}_{l=1}^L$ is then defined as:
\begin{equation}
\tilde{t}_l =
\begin{cases}
\text{[MASK]}, & l \in \mathcal{L},\\
t_l, & l \notin \mathcal{L},
\end{cases}
\end{equation}
where \text{[MASK]} denotes the special mask token \cite{devlin2019bert}.

\subsubsection{Multi-Modal Masked Reconstruction.}
Given the masked inputs $\tilde{I}$, $\tilde{P}$, and $\tilde{T}$ obtained from the previous stage, the model first extracts modality-specific representations to encode visual, geometric, and textual information. 
{Specifically, the masked image $\tilde{I}$ is encoded by the visual encoder $\phi_I$ to produce $F_I \in \mathbb{R}^{H_I \times W_I \times D_I}$. 
The masked point cloud $\tilde{P}$ is encoded by the geometric encoder $\phi_P$ to produce $F_P \in \mathbb{R}^{N_P \times D_P}$. 
The masked textual description $\tilde{T}$ is encoded by the textual encoder $\phi_T$ to produce $F_T \in \mathbb{R}^{L_T \times D_T}$.}
These representations serve as the foundational features for subsequent cross-modal embedding and reconstruction.

\par  
Visual and textual features are first fused through cross-attention 
{since} text-guided visual features help produce more accurate point cloud sampling. For visual feature ${{F}_I}$ and textual feature ${{F}_T}$, the attention is formulated as:
\begin{equation}
\begin{aligned}
\mathbf{Q}_I = {F}_I \mathbf{W}_Q^I, \quad \mathbf{K}_T = &{F}_T \mathbf{W}_K^T, \quad \mathbf{V}_T = {F}_T \mathbf{W}_V^T, \\
{F}_{I \leftarrow T} = \text{Softmax}&\Big(\frac{\mathbf{Q}_I \mathbf{K}_T^\top}{\sqrt{d}}\Big) \mathbf{V}_T,
\end{aligned}
\end{equation}
with ${F}_{T \leftarrow I}$ obtained through the same formulation as above. 
Reconstruction heads $\psi_I$ and $\psi_T$ predict the recovered image $I_{pred}$ and textual description $T_{pred}$:
 \begin{equation}
     \ I_{pred} = \psi_I({F}_{I \leftarrow T}), \quad \ T_{pred} = \psi_T({F}_{T \leftarrow I}).
 \end{equation}

The semantically enhanced visual feature ${F}_{I \leftarrow T}$ is fused with the hand point cloud feature ${F}_P^h$ via cross-attention 
{to produce} the fused enhanced hand point cloud feature ${F}_{P \leftarrow T \leftarrow I}^h$. The 
{enhanced features of the hand point cloud} ${F}_{P \leftarrow I \leftarrow T}^h$ is directly fed into a reconstruction head $\psi_P$ to predict the reconstructed hand point cloud:
\begin{equation}
P^{h}_{\text{pred}} = \psi_P({F}_{P \leftarrow I \leftarrow S}^h).
\end{equation}
The object point cloud $P^o_{\text{pred}}$ is reconstructed in the same manner. 

{This hierarchical fusion and reconstruction mechanism lets each modality draw on complementary signals from the others. It supports occlusion-aware completion and preserves semantic coherence and structural consistency across visual, geometric, and textual modalities.}

\subsection{Hand Prior Guided Pose Estimation}
To enhance the generalization capability of the model across diverse objects and interaction scenarios, we introduce a hand prior guided hand-object pose estimation module. This module leverages the inherent structural regularities of the hand as reliable prior information to guide object pose reasoning. Compared to diverse and deformable objects, the hand exhibits a relatively fixed motion topology and limited shape variation. The estimated hand pose and geometric information thus provide stable and effective spatial constraints for inferring object pose.

\subsubsection{Hand Pose Estimation.}
We incorporate an implicit geometric by predicting the signed distance field (SDF). Given hand points $P_h$, we apply Fourier positional encoding $\gamma(\cdot)$ to enrich spatial frequency representation, and fuse it with ${F}_{I \leftarrow T}$ to regress the $SDF_h$:
\begin{equation}
SDF_h = \text{MLP}_{\text{SDF}} \left( \gamma(P_h) \oplus {F}_{I \leftarrow T} \oplus P_h \right).
\end{equation}
\par The process of estimating object $SDF_o$ follows the same steps as in $SDF_h$.
We estimate the hand pose by integrating multiple cues, including the explicit geometric feature, the semantically enhanced visual feature, and the implicit geometric from the predicted $SDF_h$. The SDF-guided aggregation encourages features to focus on the hand-object surface.
\begin{equation}
F_{\text{agg}}^{h} = \left({{F}_{P \leftarrow I \leftarrow S}^h} \oplus {F}_{I \leftarrow T} \right) \cdot \frac{1}{\beta_h} \cdot \sigma_h\left(\frac{SDF_{\text{h}}}{\beta_h}\right), 
\end{equation}
\begin{equation}
(\{\theta_i\}_{i=1}^{16}, \alpha) = \text{MLP}_{h}\Big(\text{Transformer}_{h}(F_{\text{agg}}^{h})\Big),
\end{equation}
where $\sigma_h(\cdot)$ is the sigmoid function, $\beta_h$ is a learnable scale, $\theta \in \mathbb{R}^{16 \times 3}$ represents joint rotations, and $\alpha \in \mathbb{R}^{10}$ denotes MANO hand shape coefficients. The aggregation object feature ${F}_{\text{agg}}^{o}$ is obtained in the same manner as for the hand.

\subsubsection{Object Pose Estimation.}
The object rotation and translation are jointly regressed by incorporating hand cues. Specifically, the joint rotations $\theta$ are projected and globally aggregated to produce a compact hand motion descriptor, and the reconstructed 3D hand joints $J \in \mathbb{R}^{21 \times 3}$ provide spatial cues of the grasp. These hand features are concatenated with the object feature ${F}_{\text{agg}}^{o}$ to estimate the full object pose:
\begin{equation}
\begin{aligned}
R_{\text{obj}} &= \text{MLP}_R \Big({F}_{\text{agg}}^{o}\oplus \text{Pool}\big(\phi_{\text{rot}}({\theta_i}_{i=1}^{16})\big)\Big),\\
T_{\text{obj}} &= \text{MLP}_T \Big({F}_{\text{agg}}^{o} \oplus \text{Pool}\big(\phi_{\text{pos}}(J)\big) \Big).
\end{aligned}
\end{equation}
The final object pose is represented as $[R_{\text{obj}}, T_{\text{obj}}]$, where $R_{\text{obj}} \in \mathbb{R}^{3\times3}$ is a 3D rotation matrix and $T_{\text{obj}} \in \mathbb{R}^{3}$ is a 3D translation vector.

\subsection{Loss Functions}

Our overall training objective:
\begin{equation}
\begin{aligned}
L_{\text{total}} =\;& \lambda_{1} L_{\text{rec}} + \lambda_{2} L_{\text{text}} + \lambda_{3} L_{\text{pc}} \\
& + \lambda_{4} L_{\text{mano}} + \lambda_{5} L_{\text{obj}} 
+ \lambda_{6} L_{\text{SDF}} + \lambda_{7} L_{\text{others}}.
\end{aligned}
\end{equation}
combines multiple loss terms to jointly optimize multi-modal reconstruction quality and pose estimation accuracy. 
{Specifically, we implement the image reconstruction loss $L_{\text{rec}}$, the point cloud reconstruction loss $L_{\text{pc}}$, the SDF supervision loss $L_{\text{sdf}}$, the hand pose loss $L_{\text{mano}}$, and the object pose loss $L_{\text{obj}}$ as L1 losses. These loss terms are supervised by image cues, geometric consistency, implicit geometry, hand pose, and object pose, respectively.}
In contrast, the text reconstruction loss $L_{\text{text}}$ is formulated as a cross-entropy loss to reconstruct masked text tokens. The auxiliary loss terms $L_{\text{others}}$ follow Qi \textit{et al}~\cite{qi2024hoisdf} to provide task-specific constraints.
Each term is weighted by its corresponding coefficient $\lambda$ to balance its contribution during optimization.

\section{Experiments}
Our method is implemented in PyTorch and trained on a single NVIDIA RTX 4090 GPU. We adopt the Adam optimizer with a batch size of 24. The initial learning rate is set to 1e-4 and decayed by a factor of 0.7 every 5 epochs. The model is trained for 60 epochs on both the DexYCB \cite{chao2021dexycb} and HO3Dv2 \cite{hampali2020honnotate} datasets, which is sufficient to achieve satisfactory performance. Further details are in supplementary material. 


\smallskip
\noindent \textbf{Datasets.} 
We evaluate our method on two widely used hand-object interaction datasets. \textbf{DexYCB} \cite{chao2021dexycb} contains 582K images from over 1000 sequences, where 10 subjects manipulate 20 YCB objects \cite{xiang2017posecnn}. 
{We follow the S0 split for standard evaluation. The split uses sequences from 8 subjects for training and 2 subjects for testing, which keeps the test subjects and interaction patterns unseen during training. We also adopt the S3 split to evaluate generalization to unseen objects. The split uses 15 objects for training and holds out the remaining 3 objects for testing. \textbf{HO3Dv2} \cite{hampali2020honnotate} contains 77K images from 68 sequences. The dataset covers 10 subjects interacting with 10 YCB objects.}
We use the official train/test split, where the test set contains 1 unseen object category to evaluate cross-category generalization, and submit test results to the official evaluation server.

\smallskip
\noindent \textbf{Evaluation Metrics.}
\textbf{Hand Pose and Mesh Reconstruction:}
To assess hand pose accuracy, we report Mean Joint Error (M-JE), Procrustes-Aligned MJE (PA-MJE) \cite{PAMJE}, and Scale-Translation aligned MJE (ST-MJE) \cite{stnje}. Since our method directly regresses MANO parameters, we further evaluate hand mesh reconstruction quality using Mean Mesh Error (MME), the area under the curve of the percentage of correct vertices (V-AUC), and F-scores at 5mm and 15mm thresholds (F@5, F@15), together with their Procrustes-aligned versions following H2ONet \cite{xu2023h2onet}. \textbf{Object Pose:} 
For object pose estimation, we report Object Center Error (OCE), Mean Corner Error (MCE), Object Mesh Error (OME), and average closest point distance (ADD-S). We further evaluate the average 3D distance (ADD) of the object. Specifically, we report ADD-0.5D, which measures the percentage of predictions whose ADD is within 50\% of the object diameter.

\begin{table*}[t!]
\centering
\footnotesize
\caption{Evaluation of hand and object pose estimation on the DexYCB S3 split \cite{chao2021dexycb}. Results are for unseen objects and novel hand-object interactions. `avg' denotes average results over all evaluated objects. $\downarrow$ means lower is better, $\uparrow$ indicates higher is better. Best results are bolded, and the second best are \underline{underlined}. HPE: Hand Pose Estimation. HOPE: Hand-Object Pose Estimation.}
\resizebox{\textwidth}{!}{
\begin{tabular}{c|l|cccc|cccc}
\toprule[1.5pt]

\multicolumn{2}{c|}{\multirow{4}{*}{\textbf{Method}}} &
\multicolumn{4}{c|}{\textbf{Object Pose (mAP)}} &
\multicolumn{4}{c}{\textbf{Hand Pose (mm)}} \\

\cmidrule(lr){3-6} \cmidrule(lr){7-10}

\multicolumn{2}{c|}{} &
\multicolumn{4}{c|}{AUC of ADD-0.5D $\uparrow$} &
\multirow{2}{*}{\centering MJE $\downarrow$} & 
\multirow{2}{*}{\centering PA-MJE $\downarrow$} & 
\multirow{2}{*}{\centering V-PE $\downarrow$} & 
\multirow{2}{*}{\centering PA-V-PE $\downarrow$} \\

\cmidrule(lr){3-6}
\multicolumn{2}{c|}{} & gelatin\_box & bleach\_cleanser & wood\_block & avg & & & & \\
\midrule
\multicolumn{1}{c|}{\multirow{4}{*}{\rotatebox{90}{HPE}}}
 & HandOccNet \cite{park2022handoccnet} & -- & -- & -- & -- & 14.58 & 6.73 & 14.10 & 6.49 \\
\multicolumn{1}{c|}{} & MobRecon \cite{chen2022mobrecon} & -- & -- & -- & -- & 15.40 & 7.25 & 14.24 & 6.39 \\
\multicolumn{1}{c|}{} & H2ONet \cite{xu2023h2onet} & -- & -- & -- & -- & 15.20 & \underline{6.35} & 15.03 & 6.74 \\
\multicolumn{1}{c|}{} & SimpleHand \cite{zhou2024simple} & -- & -- & -- & -- & 14.88 & 6.74 & 14.21 & 6.45 \\
\midrule
\multicolumn{1}{c|}{\multirow{5}{*}{\rotatebox{90}{HOPE}}} 
 & Liu et al. \cite{liu2021semi} & \underline{26.31} & 25.07 & 68.56 & 38.89 & 15.43 & 6.61 & 14.91 & 6.40 \\
\multicolumn{1}{c|}{} & Keypoint Trans. \cite{hampali2022keypoint_transformer} & 0.00 & 1.31 & 32.61 & 10.47 & 18.79 & 7.77 & 18.35 & 7.94 \\
\multicolumn{1}{c|}{} & HFL-Net \cite{lin2023harmonious} & 25.88 & 32.08 & 70.16 & 41.59 & 14.77 & 6.64 & 14.29 & 6.41 \\
\multicolumn{1}{c|}{} & UniHOPE \cite{wang2025unihope} & 26.23 & \underline{32.32} & \underline{74.29} & \underline{43.06} & \textbf{13.31} & \textbf{6.08} & \textbf{12.89} & \textbf{5.87} \\
\multicolumn{1}{c|}{} & \textbf{GenHOI (ours)} & \textbf{86.70} & \textbf{89.27} & \textbf{92.06} & \textbf{89.34} & \underline{13.67} & 6.39 & \underline{13.32} & \underline{5.93} \\
\bottomrule[1.5pt]
\end{tabular}
}
\label{tab1:DexYCB S3 Split Quantitative Comparison}
\end{table*}

\begin{table}[t]
\centering
\renewcommand\arraystretch{1.08}
\scriptsize
\caption{Evaluation of object pose estimation using ADD-S on DexYCB S3 split \cite{chao2021dexycb}.}
\begin{tabular*}{\linewidth}{@{\extracolsep{\fill}}lcccc @{}}
\toprule[1.5pt]
\multirow{2}{*}{Method}
& \multicolumn{4}{c}{ADD-S (mm) $\downarrow$} \\
\cline{2-5}
& gelatin\_box 
& bleach\_cleanser
& wood\_block
& average
\\ 
\hline
HFL-Net \cite{lin2023harmonious}
& 94.74 
& 176.84
& 72.93 
& 114.83 
\\
UniHOPE \cite{wang2025unihope}
& 109.23
& 173.38
& \underline{49.76}
& 110.79 
 \\
HOISDF \cite{qi2024hoisdf}
& \underline{52.31}
& \underline{89.92}
& 109.89 
& {84.04} 
\\
GenHOI (ours) 
& \textbf{13.14}
& \textbf{26.91}
& \textbf{25.23} 
& \textbf{21.76} 
\\ 
\bottomrule[1.5pt]

\end{tabular*}
\label{tab2:ADD-S DexYCB S3 Split Quantitative Comparison}
\end{table}

\begin{table}[t!]
\centering
\scriptsize
\renewcommand\arraystretch{1.25}
\caption{Quantitative comparison of hand mesh with across occlusion levels on DexYCB S3 split dataset \cite{chao2021dexycb}.}
\resizebox{\linewidth}{!}{
\begin{tabular}{l|cccc|cccc|cccc}
\toprule[1.5pt]
\multirow{2}{*}{Methods} & \multicolumn{4}{c|}{Occlusion (25\%-50\%)} & \multicolumn{4}{c|}{Occlusion (50\%-75\%)} & \multicolumn{4}{c}{Occlusion (75\%-100\%)} \\ \cline{2-13} 
 & J-PE$\downarrow$ & PA-J-PE$\downarrow$ & V-PE$\downarrow$ & PA-V$\downarrow$ & J-PE$\downarrow$ & PA-J-PE$\downarrow$ & V-PE$\downarrow$ & PA-V$\downarrow$ & J-PE$\downarrow$ & PA-J-PE$\downarrow$ & V-PE$\downarrow$ & PA-V$\downarrow$ \\ \hline
HFL-Net \cite{lin2023harmonious} & 16.33 & 7.00 & 15.81 & 6.77 & 18.66 & 7.33 & 18.11 & 7.11 & 28.95 & 8.80 & 27.94 & 8.53 \\ \hline
UniHOPE \cite{wang2025unihope} & \textbf{14.59} & \textbf{6.39} & \textbf{14.13} & \textbf{6.17} & 16.27 & \textbf{6.51} & \textbf{15.78} & 6.29 & 26.42 & 7.64 & \textbf{25.51} & \textbf{7.40} \\ \hline
Ours & \underline{14.75} & \underline{6.49} & \underline{14.36} & \underline{6.23} & \textbf{16.18} & \underline{6.61} & \underline{15.89} & \textbf{6.25} & \textbf{26.28} & \textbf{7.55} & \underline{25.64} & \underline{7.57} \\ 
\bottomrule[1.5pt]
\end{tabular}
}
\label{tab3:DexYCB S3 split Occlusion Levels Comparison}
\end{table}

\subsection{Comparison with SOTA Methods}

\begin{figure}[t!]
\centering
\includegraphics[width=1\linewidth]{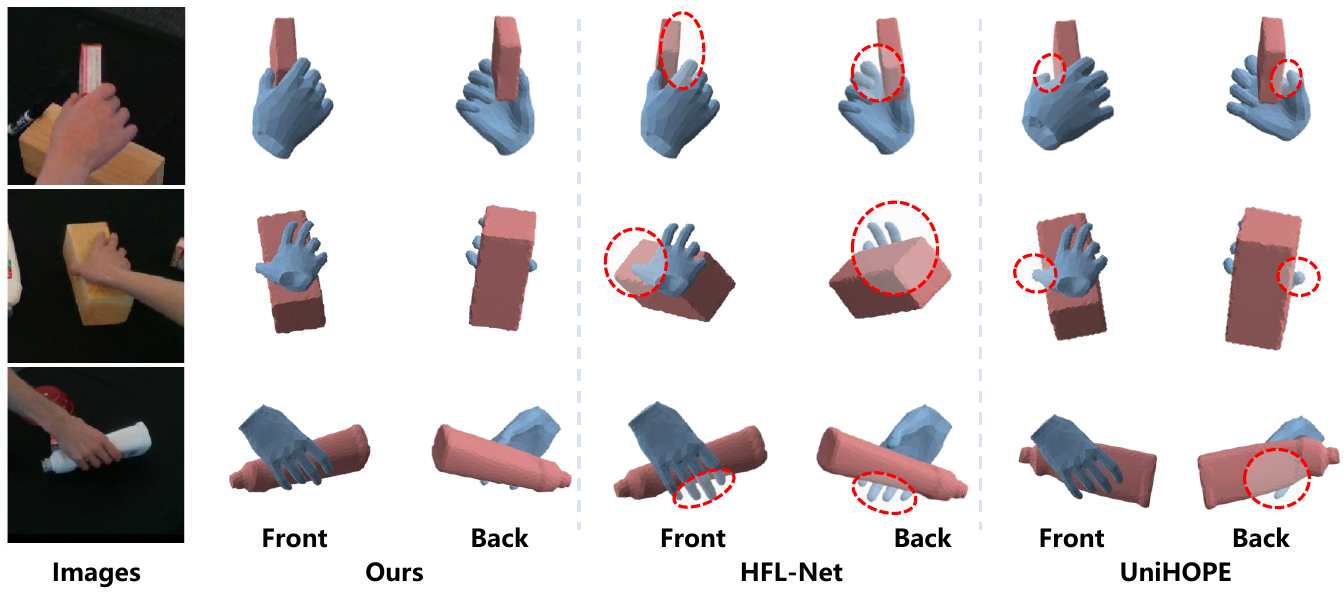}%
\caption{
Qualitative comparison of hand-object pose estimation on unseen objects under the DexYCB S3 split \cite{chao2021dexycb}. Front and back indicate the front and rear views, respectively. 
Red dotted circles highlight regions where other methods produce less accurate pose estimates than our method. 
 {This demonstrates} the superior generalization and robustness of our method on unseen objects.
}
\label{fig3:visualization of DexYCB S3 Split}
\end{figure}

\noindent \textbf{Comparisons on DexYCB.} We evaluate our method on the DexYCB S3 split dataset \cite{chao2021dexycb}, which focuses on unseen objects. Following UniHOPE \cite{wang2025unihope}, we evaluate only hand-object interaction frames. This split presents significant challenges due to severe hand-object occlusions and the inclusion of unseen objects, providing a rigorous evaluation of the generalization ability and robustness. Table \ref{tab1:DexYCB S3 Split Quantitative Comparison} presents quantitative results for hand and object pose estimation. Our method achieves competitive hand pose accuracy, closely matching state-of-the-art (SOTA) approaches. More importantly, it substantially outperforms existing methods in object pose estimation {by} achieving an average AUC of 89.34\% across all evaluated categories. 
The relatively smaller improvement in hand pose accuracy compared to object pose stems from the inherent differences in variability. The hand exhibits {a} consistent structure and appearance across samples, 
{which makes} it less influenced by the generalization design of our framework. To enable a clearer comparison under a shared mm-based object metric, we further report the ADD-S results in Table~\ref{tab2:ADD-S DexYCB S3 Split Quantitative Comparison}. The results show that our method achieves the lowest ADD-S error among all compared methods, reducing the error from 84.04 mm of HOISDF to 21.76 mm. 
{In contrast, objects exhibit much larger shape and texture variation. Our hierarchical textual semantics and hand priors therefore provide stronger semantic and geometric constraints. Following UniHOPE, we evaluate hand pose accuracy across occlusion ratios under the same protocol in Table.~\ref{tab3:DexYCB S3 split Occlusion Levels Comparison}. Our method achieves comparable performance under heavy occlusion despite UniHOPE includes a dedicated hand-occlusion removal module.}
 {Fig.~\ref{fig3:visualization of DexYCB S3 Split} shows the corresponding qualitative comparison on unseen objects under the S3 split. The results highlight improved hand-object pose accuracy over HFL-Net and UniHOPE.}

\begin{table}[t!]
\renewcommand\arraystretch{1.08}
\scriptsize
\centering
\caption{Quantitative results on the unseen object (“019 pitcher base”) from the HO3Dv2 dataset \cite{hampali2020honnotate}.}

\begin{tabular*}{\linewidth}{@{\extracolsep{\fill}} cccc @{}}
\toprule[1.5pt]
\multirow{2}{*}{\begin{tabular}[c]{@{}c@{}}Method\\ Metrics in {[}mm{]}\end{tabular}}
& \multicolumn{1}{c}{HFL-Net \cite{lin2023harmonious}}
& \multicolumn{1}{c}{HOISDF \cite{qi2024hoisdf}}
& \multicolumn{1}{c}{Ours} \\ \cline{2-4}
& ADD-0.5D & ADD-0.5D & ADD-0.5D \\ \midrule
019 pitcher base & 24.1 & 88.4 & \textbf{92.7} \\
\bottomrule[1.5pt]
\end{tabular*}
\label{tab4: HO3D Unseen Object Quantitative Comparison}
\end{table}

\begin{table}[t!]
\renewcommand\arraystretch{1.25}
\centering
\setlength{\tabcolsep}{2.5pt}
\scriptsize
\caption{Comparison with SOTA hand-object pose estimation methods on the HO3Dv2 dataset \cite{hampali2020honnotate}.}
\begin{tabular*}{\linewidth}{@{\extracolsep{\fill}} l|ccc|cc @{}}
\toprule[1.5pt]
\multicolumn{1}{c|}{Metric in {[}mm{]}} & MJE $\downarrow$ & ST-MJE $\downarrow$ & PA-MJE $\downarrow$ & OME $\downarrow$ & ADD-S $\downarrow$ \\ \midrule
Hasson \emph{et al.} \cite{hasson2019learning} & - & 31.8 & 11.0 & - & - \\
Hasson \emph{et al.} \cite{hasson20_handobjectconsist} & - & 36.9 & 11.4 & 67.0 & 22.0 \\
Hasson \emph{et al.} \cite{hasson2021towards} & - & 26.8 & 12.0 & 80.0 & 40.0 \\
Liu \emph{et al.} \cite{liu2021semi} & - & 31.7 & 10.1 & - & - \\
Hampali \emph{et al.} \cite{hampali2022keypoint_transformer} & 25.5 & 25.7 & 10.8 & 68.0 & 21.4 \\
HFL-Net \cite{lin2023harmonious} & 28.9 & 28.4 & \textbf{8.9} & 64.3 & 32.4 \\
HOISDF \cite{qi2024hoisdf} & \underline{23.6} & 22.8 & \underline{9.6} & \underline{48.5} & \underline{17.8} \\
LCP \cite{kuang2024learning} & - & 23.7 & \textbf{8.9} & - & - \\
UniHOPE \cite{wang2025unihope} & - & 25.5 & \underline{9.6} & - & - \\
\midrule
\textbf{GenHOI (ours)} & \textbf{21.3} & \textbf{21.1} & 9.9 & \textbf{35.9} & \textbf{13.4} \\
\bottomrule[1.5pt]
\end{tabular*}

\label{tab5: HO3D Seen Object Quantitative Comparison}
\end{table}

\smallskip
\noindent \textbf{Comparisons on HO3Dv2.}
We evaluate our method on the HO3Dv2 dataset \cite{hampali2020honnotate}, which mainly consists of objects seen during training, along with a single unseen object (“019 pitcher base”). To assess the generalization capability, we evaluate our model on the unseen object, as presented in Table~\ref{tab4: HO3D Unseen Object Quantitative Comparison}. Despite not being seen during training, our method achieves a competitive ADD-0.5D score of 92.7\% 
{that surpasses} existing approaches. This strong generalization to unseen objects benefits from the integration of hierarchical textual semantics and hand priors. Table~\ref{tab5: HO3D Seen Object Quantitative Comparison} reports quantitative comparisons on the seen objects. Our method consistently surpasses previous approaches across all metrics 
{in} achieving the lowest hand and object pose errors. These results demonstrate that the proposed model accurately captures hand–object interactions and yields robust estimation even under challenging occlusions.
\par
\begin{figure}[t]
\centering
\includegraphics[width=1\linewidth]{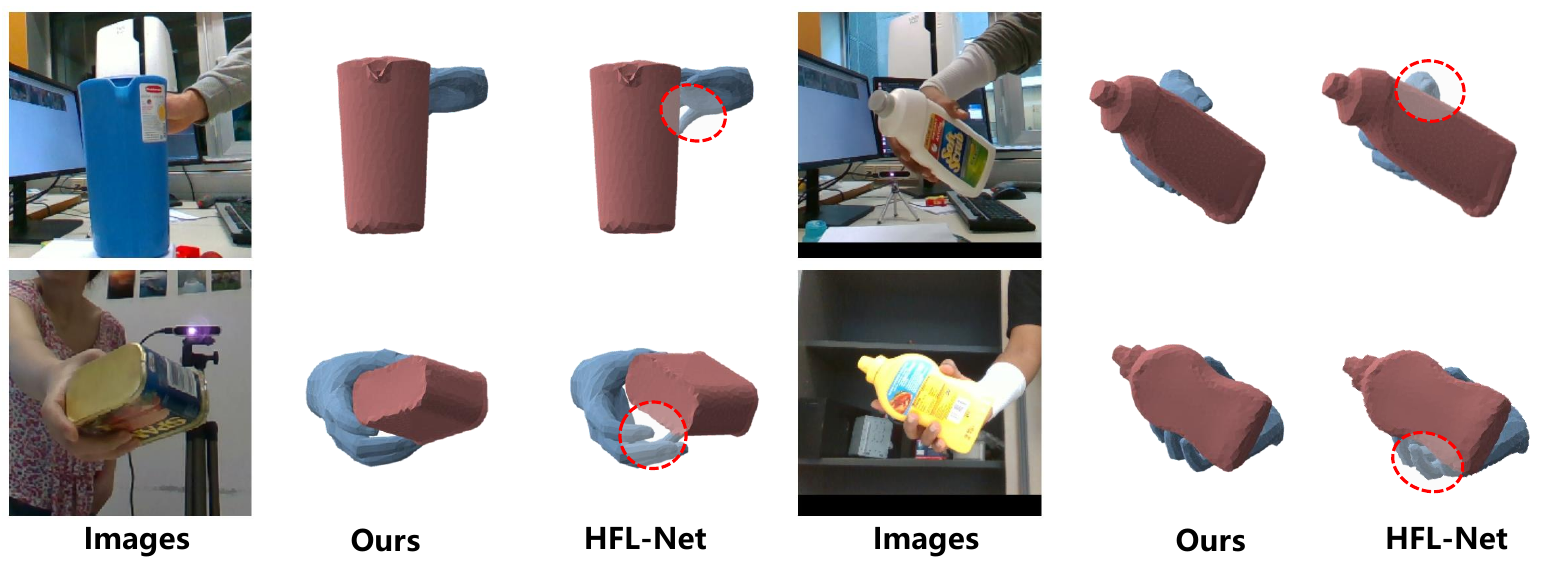}%
\caption{
Qualitative comparison on the HO3Dv2 dataset~\cite{hampali2020honnotate}.
The top-left example show results on the unseen object (“019 pitcher base”), while the remaining examples correspond to objects seen during training.
}
\label{fig4:visualization of HO3D}
\end{figure}

 Qualitative comparisons on both unseen and seen objects are shown in Fig.~\ref{fig4:visualization of HO3D}. Specifically, the top-left example corresponds to the unseen object, while the remaining examples show results on seen objects. The qualitative results further confirm that our method attains SOTA performance on HO3Dv2 while maintaining robust generalization beyond the training distribution.

\begin{table*}[t]
\renewcommand\arraystretch{0.9}
\centering
\caption{Ablation study of our proposed components on the DexYCB dataset S3 split full \cite{chao2021dexycb}. `w/o' denotes ‘without’.}
\small
\resizebox{\textwidth}{!}{
\begin{tabular}{c|lcccccc}
\toprule[1.5pt]
\multicolumn{2}{l|}{Ablation} & MJE $\downarrow$ & PA-MJE $\downarrow$ & V-PE $\downarrow$ & PA-V-PE $\downarrow$ & ADD-0.5D $\uparrow$ \\ \midrule

\multicolumn{7}{c}{\textbf{Hierarchical Textual Semantic}} \\ \midrule
1 & \multicolumn{1}{l|}{w/o Text Generation Template}      & 16.78 & 6.73 & 15.12 & 6.23 & 86.51 \\
2 & \multicolumn{1}{l|}{w/o Multi-level Semantic Knowledge} & 17.04 & 6.98 & 15.75 & 6.41 & 86.32 \\ \midrule

\multicolumn{7}{c}{\textbf{Multi-Modal Masked Modeling}} \\ \midrule
3 & \multicolumn{1}{l|}{w/o Image Masking}                  & 15.33 & 6.52 & 14.78 & 6.03 & 87.65 \\
4 & \multicolumn{1}{l|}{w/o Text Masking}                   & 14.98 & 6.21 & 14.50 & 5.88 & 88.12 \\
5 & \multicolumn{1}{l|}{w/o Point Cloud Masking}            & 15.21 & 6.47 & 14.91 & 6.07 & 87.90 \\ \midrule

\multicolumn{7}{c}{\textbf{Hand Priors}} \\ \midrule
6 & \multicolumn{1}{l|}{w/o Hand Rotation Prior}            & 14.65 & 6.12 & 14.32 & 5.78 & 85.45 \\
7 & \multicolumn{1}{l|}{w/o Hand Joint Prior}               & 14.88 & 6.29 & 14.55 & 5.91 & 84.01 \\ \midrule

 \multicolumn{2}{l|}{\textbf{GenHOI (Full)}}             & \textbf{13.42} & \textbf{5.81} & \textbf{12.96} & \textbf{5.63} & \textbf{90.44} \\ 
\bottomrule[1.5pt]
\end{tabular}}
\label{tab6:my_ablation}
\end{table*}

\subsection{Ablation Studies}

We perform the ablations on the DexYCB S3 split full \cite{chao2021dexycb}, using all frames to evaluate the contribution of each component.

\noindent \textbf{Hierarchical Textual Semantic.} Rows (1-2) of Table \ref{tab6:my_ablation} show apparent performance drops when removing either the text template or hierarchical semantic descriptions. Our motivation for introducing hierarchical semantics is to provide structured level-dependent priors about object and interaction attributes. These cues offer high-level concepts as well as fine-grained part-level hints, 
{which guide} the model to reason beyond purely visual correlation. 
{The model is forced to rely mainly on appearance cues without them, and thus weakening its generalizability.}
The resulting performance gap confirms that hierarchical semantics serve as a semantic generalization prior, 
{which is} crucial when handling novel hand-object interaction scenarios with severe occlusion.

\smallskip
\noindent \textbf{Multi-Modal Masked Modeling.} Rows (3-5) of Table \ref{tab6:my_ablation} indicate that image masking contributes the most to performance, 
{and} text and point cloud masking further enhance occlusion awareness. The multi-modal masking strategy enables the model to infer occluded information. 
{The model develops occlusion-aware reasoning and cross-modal feature agreement by learning to reconstruct from incomplete observations, which are crucial for robust performance in complex interaction scenes.}
The consistent performance degradation when 
{the masks are removed} validates that explicitly learning to reason under occlusion is central to improving generalization in real-world settings.

\smallskip
\noindent \textbf{Hand Priors.} Rows (6-7) of Table \ref{tab6:my_ablation} show that 
{the removal of} hand priors only slightly affects hand pose estimation but leads to a clear drop in object pose accuracy, 
{which indicates} that stable hand priors provide essential spatial constraints for interaction reasoning. This is because hand configuration and articulation inherently encode grasp intent and contact topology, which implicitly restrict the feasible orientation and position of the object in 3D space. Even when the hand pose can still be estimated from appearance cues, the loss of these structured priors weakens the geometric coupling between hand and object. As a result, object pose inference becomes more sensitive to local visual variations. 
{The model relies more on appearance similarity without such constraints. This makes object pose inference harder when shapes and interaction patterns vary, which reduces generalization to unseen objects and novel interaction scenarios.}

\smallskip
\noindent \textbf{Additional ablations and analyses} on multimodal mask ratios, multimodal fusion strategies, robustness to inaccurate or noisy text descriptions, as well as inference efficiency are provided in the supplementary material.

\section{Conclusion}

In this work, we presented GenHOI, a novel hand-object pose estimation framework that improves generalization to unseen objects and interaction scenarios. We introduced hierarchical textual semantics to encode object states, hand configurations, and interaction patterns 
{that provide} high-level guidance beyond visual appearance. We proposed a multi-modal masked modeling strategy that enables the model to infer occluded regions and maintain cross-modal consistency under partial observations. Furthermore, hand priors serve as stable spatial constraints that anchor interaction geometry 
{to reinforce} object pose reasoning even when object shapes vary. Extensive experiments demonstrate that {our} GenHOI achieves strong generalization and robust performance under severe occlusion. 
{For future work, we may extend the model to egocentric datasets and further validate its generalization.}

\section*{Acknowledgements}
This work is supported by the National Natural Science Foundation of China under Grants 62473141 and U22A2059, National Science and Technology Major Project of China under Grants 2026ZD1610900, China Scholarship Council under Grant 202506130069, the Open Foundation of the State Key Laboratory of Advanced Design and Manufacturing for Vehicle Body, the National Research Foundation (NRF) Singapore, under
its NRF-Investigatorship Programme (Award ID. NRF-NRFI09-0008), and the Tier 2 grant MOET2EP20124-0015 from the Singapore Ministry of Education.

\bibliographystyle{splncs04}
\bibliography{main}

\clearpage
\section*{Appendix}
In this supplementary material, we first introduce the implementation details (Section A). Section B presents additional qualitative results and ablation experiments to further demonstrate the robustness of GenHOI, and Section C discusses the limitations and future directions.

\section*{A. Implementation Details}
The input images are cropped around the object and resized to 224×224. In our proposed masking strategy, the number of image masking blocks is denoted as \(N_I\) and set to 12. For masks falling within the object bounding box, the parameter \(\alpha\) is configured to 50\%. For text masking, the masking ratio is denoted as \(\gamma\) and set to 50\%. For point cloud masking, the number of groups is denoted as \(K\) and set to 64, while the proportion of point cloud masking groups corresponding to the parameter \(\beta\) is 30\%. The number of hand points is denoted as \(N_H\) and set to 600, and the number of object points is denoted as \(N_O\) and set to 200.
The feature dimensions for images, point clouds and text are respectively denoted as \(D_I\), \(D_P\) and \(D_T\). These three feature dimensions are set to 768, 768 and 512. Through extensive experimental validation, it has been found that setting \(\lambda_1\), \(\lambda_2\), \(\lambda_3\), \(\lambda_4\), \(\lambda_5\) and \(\lambda_6\) to 1 achieves the optimal performance. The parameter \(\lambda_7\) is configured according to the parameter settings specified in HOISDF \cite{qi2024hoisdf}.

\section*{B. More Experimental Results}

\section*{B.1 Quantitative Results on the DexYCB S0 Split}

We report quantitative results on the DexYCB S0 split dataset \cite{chao2021dexycb} in Tables~\ref{tab3:tab1:DexYCB S0 Split Quantitative Comparison} and~\ref{tab4:DexYCB S0 Split Hand Mesh Quantitative Comparison}.
We evaluate our method on the standard S0 split of the DexYCB dataset \cite{chao2021dexycb}, where the test set contains objects seen during training. 
Table \ref{tab3:tab1:DexYCB S0 Split Quantitative Comparison} summarizes the quantitative comparison with SOTA methods. Our approach achieves the best performance across all reported metrics.
The corresponding qualitative results for the S0 split are provided in the supplementary material.

\par In addition, as our method involves MANO-based hand reconstruction, we further compare it with SOTA hand mesh reconstruction methods in Table \ref{tab4:DexYCB S0 Split Hand Mesh Quantitative Comparison}. Our method achieves the best performance under both procrustes-aligned and non-aligned metrics, demonstrating its joint-level accuracy and detailed mesh reconstruction capability under complex occlusions and hand-object interactions.

\section*{B.2 Qualitative Results on the DexYCB S0 Split}
\begin{figure*}[t!]
\renewcommand{\thefigure}{A} 
\centering
\includegraphics[width=1\linewidth]{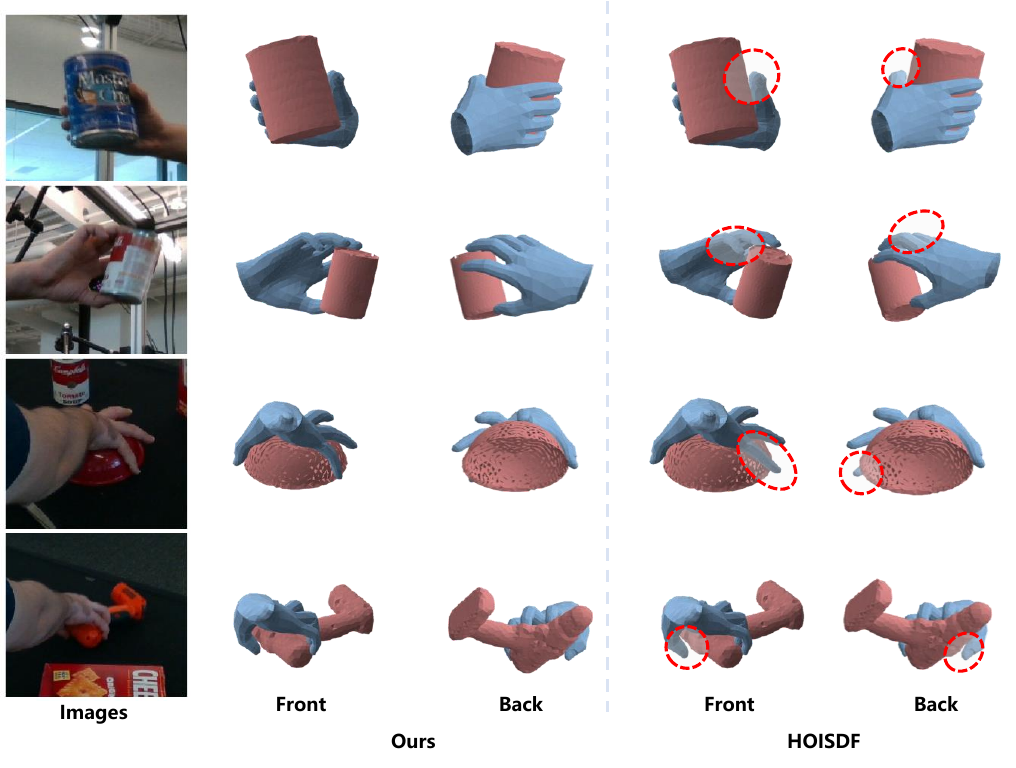}%
\vspace{-1em}
\caption{
Qualitative comparison between our method and HOISDF \cite{qi2024hoisdf} on the DexYCB S0 split \cite{chao2021dexycb}.
}
\label{fig:comparison with HOISDF on the DexYCB S0 split}
\end{figure*}

\begin{figure*}[t!]
\renewcommand{\thefigure}{B} 
\centering
\includegraphics[width=0.95\linewidth]{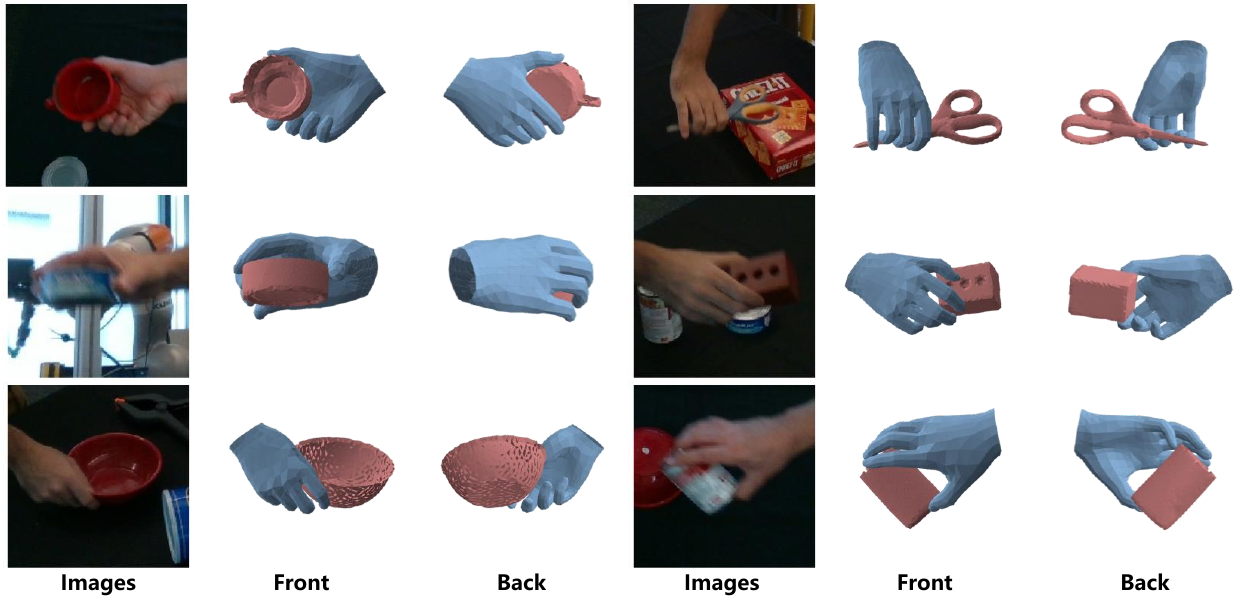}%
\vspace{-0.5em}
\caption{
Qualitative results of our method on the DexYCB S0 split \cite{chao2021dexycb}.
}
\label{fig:visualization on the DexYCB S0 split}
\vspace{-1em}
\end{figure*}

We demonstrate the superiority of our method on the DexYCB S0 split dataset \cite{chao2021dexycb} in Figs. \ref{fig:comparison with HOISDF on the DexYCB S0 split} and Figs. \ref{fig:visualization on the DexYCB S0 split}.

\begin{table}[t!]
\renewcommand{\thetable}{A}
\renewcommand\arraystretch{1.25}
\centering
\scriptsize
\caption{Comparison with SOTA hand-object pose estimation methods on the DexYCB S0 split dataset \cite{chao2021dexycb}.}

\setlength{\tabcolsep}{3pt}
\begin{tabular*}{\linewidth}{@{\extracolsep{\fill}} l|cc|ccc @{}}
\toprule[1.5pt]
\multicolumn{1}{c|}{Metric in {[}mm{]}} & MJE $\downarrow$ & PA-MJE $\downarrow$& OCE $\downarrow$  & MCE $\downarrow$ & ADD-S $\downarrow$ \\ \midrule
Hasson \emph{et al.} \cite{hasson2019learning}      & 17.6 & -     & -    & -    & -       \\
Hasson \emph{et al.} \cite{hasson2021towards}      & 18.8 & -     & -    & 52.5 & -        \\
Tze \emph{et al.} \cite{tze}          & 15.3 & -     & -    & -    & -      \\
ArtiBoost \cite{yang2022artiboost}           & 12.8 & -     & -    & -    & -         \\
AlignSDF  \cite{chen2022alignsdf}         & 19.0 & -     & 27.0 & -    & -         \\
gSDF \cite{chen2023gsdf}         & 14.4 & -     & 19.1 & -    & -        \\
Keypoint Trans. \cite{hampali2022keypoint_transformer}        & 12.7 & 6.86  & 27.3 & 32.6 & 15.9      \\
HFL-Net \cite{lin2023harmonious}         & 11.9 & 5.81  & 39.8 & 45.7 & 31.9    \\
LCP \cite{kuang2024learning}          & 11.1 & 5.30  & -    & -    & -    \\ 
HOISDF \cite{qi2024hoisdf}          & \underline{10.1} & \underline{5.31}  & \underline{18.4} & \underline{27.4} & \underline{13.3}     \\
UniHOPE \cite{wang2025unihope}  & 12.6  & 5.54  & -    & -    & - \\  \midrule
\textbf{GenHOI (ours)}               & \textbf{9.83} & \textbf{5.12} & \textbf{17.2} & \textbf{25.1} & \textbf{12.3} \\
\bottomrule[1.5pt]
\end{tabular*}

\label{tab3:tab1:DexYCB S0 Split Quantitative Comparison}
\end{table}

\begin{table*}[t!]
\renewcommand{\thetable}{B}
\renewcommand\arraystretch{1.25}
\footnotesize
   \setlength{\tabcolsep}{4.1pt}

\centering
\caption{Quantitative comparison with hand mesh metrics on the DexYCB S0 split dataset \cite{chao2021dexycb}.}
\resizebox{\textwidth}{!}{
\begin{tabular}{l|ccccc|ccccc}
\toprule[1.5pt]
Methods                                              & PA-J-AUC$\uparrow$ & PA-V-PE$\downarrow$ & PA-V-AUC$\uparrow$ & PA-F@5$\uparrow$ & PA-F@15$\uparrow$ & J-AUC$\uparrow$ & V-PE$\downarrow$ & V-AUC$\uparrow$ & F@5$\uparrow$ & F@15$\uparrow$ \\ \hline
HandOccNet\cite{park2022handoccnet} & 88.4               & 5.5                 & 89.0               & 78.0             & 99.0     & \underline{74.8}            & 13.1             & 76.6            & 51.5          & 92.4           \\
MobRecon\cite{chen2022mobrecon}     & 87.3               & 5.6                 & 88.9               & 78.5             & 98.8              & 73.7            & 13.1             & 76.1            & 50.8          & 92.1           \\
H2ONet \cite{akiva2021h2o}          & 88.9               & 5.5                 & 89.1               & 80.1             & 99.0     & 74.6            & 13.0             & 76.2            & 51.3          & 92.1           \\
HOISDF \cite{qi2024hoisdf}          & \underline{89.8}               & \underline{5.1}                 & \underline{89.8}               & \underline{80.7}            & \underline{99.2}    & \textbf{80.7}           & \underline{9.8}             & \underline{80.6}            & \underline{59.6}          & \underline{95.0}           \\
UniHOPE \cite{wang2025unihope}      & -                  & 5.4                 & -                  & -                & -                 & -               & 12.2             & -               & -             & -              \\ \hline
Ours                                                 & \textbf{90.0}      & \textbf{4.9}        & \textbf{90.1}      & \textbf{81.3}    & \textbf{99.4}              & \textbf{80.7}   & \textbf{9.5}    & \textbf{81.1}   & \textbf{60.4} & \textbf{95.4}  \\ 
\bottomrule[1.5pt]
\end{tabular}}
\label{tab4:DexYCB S0 Split Hand Mesh Quantitative Comparison}
\end{table*}

\begin{table}[t]
\renewcommand{\thetable}{C}
\centering
\footnotesize
\setlength{\tabcolsep}{3.5pt}
\renewcommand{\arraystretch}{1.15}

\newcommand{\best}[1]{\textbf{#1}}
\newcommand{\second}[1]{\underline{#1}}
\vspace{-1em}
\caption{Occlusion-aware ablation study of the proposed components on the DexYCB S3 split, following the ablation indices in Table~5.}
\label{tab:dexycb_occlusion}
\resizebox{\textwidth}{!}{
\begin{tabular}{c *{15}{c}}
\toprule
\multirowcell{2}{Ablation}
& \multicolumn{5}{c}{Occlusion 25\%--50\%}
& \multicolumn{5}{c}{Occlusion 50\%--75\%}
& \multicolumn{5}{c}{Occlusion 75\%--100\%} \\
\cmidrule(lr){2-6}
\cmidrule(lr){7-11}
\cmidrule(lr){12-16}
& \makecell{J-PE}
& \makecell{PA-J-PE}
& \makecell{V-PE}
& \makecell{PA-V}
& ADD
& \makecell{J-PE}
& \makecell{PA-J-PE}
& \makecell{V-PE}
& \makecell{PA-V}
& ADD
& \makecell{J-PE}
& \makecell{PA-J-PE}
& \makecell{V-PE}
& \makecell{PA-V}
& ADD \\
\midrule

1
& 16.87 & 6.82 & 15.62 & 6.51 & 86.20
& 18.87 & 7.06 & 17.88 & 6.93 & 84.35
& 30.45 & 8.54 & 29.38 & 8.42 & 78.70 \\

2
& 17.16 & 7.03 & 16.18 & 6.68 & 85.95
& 19.24 & 7.31 & 18.42 & 7.05 & 84.05
& 31.10 & 8.87 & 30.08 & 8.73 & 78.10 \\

3
& 15.68 & 6.72 & 15.22 & 6.38 & 87.45
& 17.24 & 6.92 & 16.88 & 6.50 & 85.90
& 28.10 & 8.05 & 27.52 & 8.02 & 80.65 \\

4
& 15.39 & 6.60 & 14.95 & 6.32 & 88.10
& 16.83 & 6.80 & 16.48 & 6.43 & 86.55
& 27.42 & 7.91 & 26.84 & 7.89 & 81.45 \\

5
& 15.58 & 6.68 & 15.14 & 6.37 & 87.72
& 17.08 & 6.87 & 16.73 & 6.49 & 86.15
& 27.90 & 8.00 & 27.26 & 7.98 & 80.95 \\

6
& 15.28 & 6.56 & 14.82 & 6.30 & 84.90
& 16.62 & 6.74 & 16.31 & 6.38 & 83.25
& 27.24 & 7.83 & 26.68 & 7.86 & 77.40 \\

7
& 15.36 & 6.62 & 14.97 & 6.34 & 83.70
& 16.75 & 6.82 & 16.45 & 6.45 & 81.95
& 27.52 & 7.95 & 26.96 & 7.94 & 75.85 \\

\bottomrule
\end{tabular}}
\end{table}

\begin{table*}[t!]
\renewcommand\arraystretch{1.25}
\renewcommand{\thetable}{D}
\footnotesize
\centering
\caption{
Ablation on the image masking ratio $\alpha$ and the number of masking patches $N_I$. 
Here, $\alpha$ controls the proportion of masked regions within the object bounding box, and $N_I$ denotes the total number of masked image patches. 
``All'' indicates that all patches inside the bounding box are masked.
}
\vspace{-0.5em}
\resizebox{\textwidth}{!}{
\begin{tabular}{cc|cccc|c}
\toprule[1.5pt] \textbf{Image Mask Ratio} &\textbf{Image Mask Pacth} & MJE $\downarrow$ & PA-MJE $\downarrow$ & V-PE $\downarrow$ & PA-V-PE $\downarrow$ & ADD-0.5D $\uparrow$ \\ \midrule

0\%   & 12 & 15.33 & 6.52 & 14.78 & 6.03 & 87.65 \\
25\%  & 12 & 14.15 & 6.21 & 13.86 & 5.94 & 88.17 \\
\textbf{50\%} & \textbf{12 (Ours)} & \textbf{13.42} & \textbf{5.81} & \textbf{12.96} & \textbf{5.63} & \textbf{90.44} \\
75\%  & 12 &15.48 & 6.43 & 14.96 & 6.31 & 87.95 \\
100\%  & 12 & 16.75 & 6.89 & 16.29 & 6.56 & 85.68 \\ \hline

50\%  & 4  & 14.98 & 6.39 & 14.63 & 6.15 & 88.32 \\
50\%  & 8  & 13.98 & 6.10 & 13.64 & 5.93 & 89.12 \\
\textbf{50\%}  & \textbf{12 (Ours)} & \textbf{13.42} & \textbf{5.81} & \textbf{12.96} & \textbf{5.63} & \textbf{90.44} \\
50\%  & 16 & 13.75 & 6.08 & 13.41 & 5.95 & 89.01 \\ \midrule
100\%  & All & 24.86 & 11.69& 25.73 & 10.95 & 73.21 \\
\bottomrule[1.5pt]
\end{tabular}}
\label{tab:ablation on image masking}
\end{table*}

\begin{table*}[t!]
\centering
\renewcommand{\thetable}{E}
\footnotesize
\caption{
Ablation study on the text masking ratio $\gamma$. 
We vary the proportion of masked tokens while keeping all other settings fixed.
}
\vspace{-0.5em}
\begin{tabular}{c|cccc|c}
\toprule[1.2pt]
\textbf{Text Masking Ratio} & MJE $\downarrow$ & PA-MJE $\downarrow$ & V-PE $\downarrow$ & PA-V-PE $\downarrow$ & ADD-0.5D $\uparrow$ \\
\midrule
0\%     & 14.98 & 6.21 & 14.50 & 5.88 & 88.12 \\
25\%     & 14.26 &5.96 & 14.13 & 5.82 & 88.75 \\

\textbf{50\% (Ours)} & \textbf{13.42} & \textbf{5.81} & \textbf{12.96} & \textbf{5.63} & \textbf{90.44} \\
75\%   & 15.79 & 6.45 & 15.13 & 6.13 & 84.05 \\
100\%   & 17.85 & 7.25 & 16.98 & 6.98 & 79.21 \\

\bottomrule[1.2pt]
\end{tabular}
\label{tab:ablation on text masking}
\end{table*}

\begin{table*}[t!]
\centering
\renewcommand{\thetable}{F}
\footnotesize
\caption{
Ablation study on the point cloud masking ratio. 
We fix the number of point cloud groups to $K=64$ and vary the proportion of groups being masked.
}
\vspace{-0.5em}
\resizebox{\textwidth}{!}{
\begin{tabular}{c|cccc|c}
\toprule[1.2pt]
\textbf{Point Cloud Masking Ratio} & MJE $\downarrow$ & PA-MJE $\downarrow$ & V-PE $\downarrow$ & PA-V-PE $\downarrow$ & ADD-0.5D $\uparrow$ \\
\midrule
0\%     & 15.21 & 6.47 & 14.91 & 6.07 & 87.90\\
\textbf{30\% (Ours)} & \textbf{13.42} & \textbf{5.81} & \textbf{12.96} & \textbf{5.63} & \textbf{90.44} \\
50\% & 17.35 & 7.71 & 17.04 & 7.28 & 82.23 \\
70\%                 & 24.89 & 9.56& 25.45 & 8.42 & 71.19 \\
100\% & 36.05 & 13.79 & 36.67 & 14.84 & 54.93 \\
\bottomrule[1.2pt]
\end{tabular}}
\label{tab:ablation on pointcloud masking}
\end{table*}

\begin{table*}[t!]
\centering
\renewcommand{\thetable}{G}
\footnotesize
\caption{Ablation study on the cross-attention order among text, image, and point cloud modalities. 
}
\vspace{-0.5em}
\resizebox{\textwidth}{!}{
\begin{tabular}{c|cccc|c}
\toprule[1.2pt]
\textbf{Fusion Order} & MJE $\downarrow$ & PA-MJE $\downarrow$ & V-PE $\downarrow$ & PA-V-PE $\downarrow$ & ADD-0.5D $\uparrow$ \\
\midrule
Text$\rightarrow$Point$\rightarrow$Image & 14.27 & 6.23 & 13.85 & 6.12 & 87.18 \\
Text$\rightarrow$Image \& Text$\rightarrow$Point & 13.76 & 6.02 & 13.24 & 5.89 & 88.76 \\
\textbf{Text$\rightarrow$Image$\rightarrow$Point (Ours)} & \textbf{13.42} & \textbf{5.81} & \textbf{12.96} & \textbf{5.63} & \textbf{90.44} \\ 
\bottomrule[1.2pt]
\end{tabular}}
\label{tab:ablation on fusion order}
\end{table*}

\section*{B.3 Additional Ablation Results}
To further analyze our model design and in accordance with the main paper, we conducted all of the following ablation studies on DexYCB S3 split \cite{chao2021dexycb}.

\vspace{0.6em}

\noindent\textbf{Ablation on Proposed Components under Different Occlusion Levels.}
As shown in Table~\ref{tab:dexycb_occlusion}, we further analyze the robustness of our proposed components under different occlusion levels on the DexYCB S3 split. The results show that performance degradation becomes more pronounced as the occlusion ratio increases, indicating the importance of these components for occlusion-aware hand-object reasoning. Hierarchical textual semantics provide high-level object and interaction guidance to compensate for missing or ambiguous visual evidence. Multi-modal masked modeling improves robustness by encouraging reasoning from complementary image, text, and point-cloud cues. Hand priors regularize plausible hand articulation and rotation when visible hand regions are incomplete. The consistent drops caused by removing these components further validate their effectiveness under occlusion.

\noindent \textbf{Ablation on Image Object-Centric Masking Ratio and the Total Number of Masking Patches.} 
As shown in Table \ref{tab:ablation on image masking}, we analyze how the image masking ratio and the number of masking patches affect model robustness. With a fixed patch count, increasing the masking ratio initially improves performance by encouraging reliance on stable global structure rather than occlusion-prone local textures. When the ratio too large, excessive   visual information loss prevents reliable structural reasoning and leads to degradation, indicating that the ratio must balance occlusion simulation with essential cue preservation. With a fixed masking ratio, a moderate number of patches performs best. Too few patches offer limited occlusion diversity, while too many fragment the masked regions and disrupt object structure. Overall, image masking strategy provides the greatest robustness gain only when the masking ratio and patch count are jointly balanced. As shown in the last row of Table~\ref{tab:ablation on image masking}, when the entire object bounding box is masked, most object-related and some hand-related visual cues are occluded, leading to a severe drop in pose estimation accuracy.

\noindent \textbf{Ablation on Text Masking Ratio.} 
To evaluate the impact of masking ratio on text semantic infusion, we conduct an ablation study by varying the text masking ratio while keeping all other settings fixed. As shown in Table \ref{tab:ablation on text masking}, compared with low masking ratios, a moderate masking ratio yields better performance because the model leverages textual contextual semantics to reconstruct missing information, enhancing its robustness and contextual understanding. Conversely, excessively high masking ratios remove too much semantic information, limiting the model to infer missing information from context and weakening the textual semantic supplement for occluded visual cues. These results indicate that an appropriate masking ratio enhances the model ability to reason over contextual semantics.

\noindent \textbf{Ablation on Point Cloud Masking Ratio.} To evaluate how the point cloud masking ratio affects geometric reasoning, we vary the proportion of masked point cloud groups while fixing the total number of groups at $K=64$. As shown in Table \ref{tab:ablation on pointcloud masking}, a moderate masking ratio of 30\% achieves the best overall performance. This level of masking provides enough geometric degradation for the model to learn occlusion-aware reasoning, encouraging it to infer missing geometry using context. In contrast, a very low masking ratio offers limited improvement because the model seldom encounters incomplete geometry and thus cannot effectively learn to handle occluded or sparse point clouds. However, when the masking ratio becomes excessively high, performance drops sharply because too many point cloud groups are removed, leaving insufficient local samples for reliable SDF learning and too little geometric structure for the model to reconstruct the missing regions from contextual cues. 
Overall, these results demonstrate that a balanced masking ratio is essential: it must be large enough to promote occlusion-aware learning, yet not so large that geometric supervision and contextual reconstruction collapse.

\noindent \textbf{Ablation on Cross-Attention Order.} Table~\ref{tab:ablation on fusion order} compares different cross-attention orders among text, image, and point cloud modalities. When the model directly fuses text with point cloud features before visual modality (\textit{Text → Point → Image}), the performance drops significantly. This is because textual tokens encode high-level semantic concepts, while point cloud features represent raw local geometry; without intermediate visual grounding, the model struggles to establish reliable correspondences between abstract semantic cues and local geometry. The parallel fusion strategy (\textit{Text → Image \& Text → Point}) alleviates this issue but still suffers from inconsistent feature correspondence between the two branches. In contrast, our sequential order (\textit{Text → Image → Point}) first injects textual semantics into the visual features to establish semantic-visual alignment and then transfers the enriched visual cues to the 3D representation, resulting in more coherent multi-modal interaction and improved estimation accuracy across all metrics.

\begin{table*}[t!]
\centering
\renewcommand{\thetable}{H}
\footnotesize
\caption{Ablation on the hierarchical template. We assess each component by retaining only the object-, hand-, or interaction-related part of the hierarchical template.}
\vspace{-0.5em}
\resizebox{\textwidth}{!}{
\begin{tabular}{c|cccc|c}
\toprule[1.2pt]
\textbf{Template Configuration} & MJE $\downarrow$ & PA-MJE $\downarrow$ & V-PE $\downarrow$ & PA-V-PE $\downarrow$ & ADD-0.5D $\uparrow$ \\
\midrule
Object Component Only        & 14.52 & 6.37 & 14.31 & 6.28 & 88.56 \\
Hand Component Only          & 13.98 & 6.13 & 13.65 & 6.06 & 87.15 \\
Interaction Component Only   & 13.82 & 6.05 & 13.36 & 5.81 & 89.27 \\
w/o YCB category tokens   & 13.45 & 5.78 & 13.03 & 5.61 & 90.22 \\
Full Template (ours)         & \textbf{13.42} & \textbf{5.81} & \textbf{12.96} & \textbf{5.63} & \textbf{90.44} \\
\bottomrule[1.2pt]
\end{tabular}}
\label{tab:ablation on text template}
\end{table*}

\begin{table*}[t!]
\centering
\renewcommand{\thetable}{I}
\footnotesize
\caption{Ablation on noise augmentation. We study the effect of adding noise to text and point cloud to improve model robustness.}
\vspace{-0.5em}
\resizebox{\textwidth}{!}{
\begin{tabular}{c|cccc|c}
\toprule[1.2pt]
\textbf{Noise Augmentation} & MJE $\downarrow$ & PA-MJE $\downarrow$ & V-PE $\downarrow$ & PA-V-PE $\downarrow$ & ADD-0.5D $\uparrow$ \\
\midrule
w/o Text Noise & 14.55 & 6.18 & 13.48 & 6.24 & 89.12 \\
w/o Point Cloud Noise & 14.41 & 6.21 & 13.82 & 6.28 & 88.53 \\
Full Noise (ours)  & \textbf{13.42} & \textbf{5.81} & \textbf{12.96} & \textbf{5.63} & \textbf{90.44} \\ 
\bottomrule[1.2pt]
\end{tabular}}
\label{tab:ablation on noise augmentation}
\end{table*}

\noindent \textbf{Ablation on Template Ablation.} Table~\ref{tab:ablation on text template} evaluates the contribution of each component in the hierarchical template.
Using only the object-related template provides global cues about category and shape but omits hand articulation and contact information, limiting its ability to resolve interaction geometry.
Conversely, the hand-only template lacks object attributes, making it inadequate for constraining object pose under occlusion.
The interaction-only template captures relational intent but excludes  hand-obejct specific details, leading to poor performance when visual cues are ambiguous.
In contrast, the full hierarchical template, which combines object, hand, and interaction semantics, consistently improves accuracy across all metrics. This demonstrates that a richer semantic context helps the model better understand the interaction and infer more precise hand-object poses.
\par In addition, we include an ablation without YCB category tokens, where object categories are inferred by the LLM rather than provided as ground-truth labels, to explicitly examine whether category information is leaked to the model. As shown in In the fourth row of Tab.~\ref{tab:ablation on text template}, removing YCB category tokens results in a marginal change in accuracy. This indicates that explicit category labels are not essential for pose estimation. Instead, the model mainly relies on high-level appearance cues and hand-object interaction semantics encoded in the hierarchical template.

\noindent \textbf{Ablation on Noise Augmentation.} Table~\ref{tab:ablation on noise augmentation} evaluates the effect of adding noise to text descriptions and point clouds. Our method uses hierarchical prompts generated by InstructBLIP \cite{dai2023instructblip}, whose quality, consistency, and potential biases may affect pose estimation. To reduce sensitivity to partially incorrect, biased, or domain-shifted textual inputs, we introduce token-level perturbations during training, allowing the model to learn robustness to imperfect text. Similarly, point cloud noise is added during training to mitigate the distribution gap between ideal training samples (from ground-truth sampling) and the voxel-based sampling used at inference. The results show that removing either type of noise degrades performance, while applying both jointly (\textit{Full Noise}) achieves the best accuracy, demonstrating that these augmentations improve robustness and generalization under  imperfect inputs.

\noindent \textbf{Runtime Analysis.} To assess the computational efficiency of our proposed framework, we evaluate the inference speed, parameter count, and memory consumption. The model achieves a inference speed of 18 FPS (text generated offline), demonstrating its practicality for real-world applications. The total parameter count remains moderate at 292 MB, and the memory footprint is limited to approximately 5109 MB. These results indicate that our framework achieves a good balance between efficiency and accuracy, making it suitable for deployment in scenarios requiring real-time or near real-time hand–object pose estimation.


\begin{figure*}
\vspace{-1em}
\renewcommand{\thefigure}{C} 
\centering
\includegraphics[width=1\linewidth]{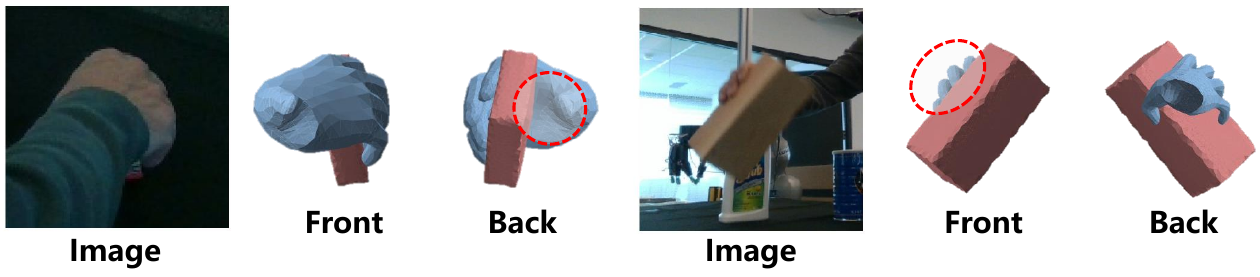}%
\vspace{-0.5em}
\caption{Failure cases. In scenarios with severe occlusions, the predicted hand-object poses exhibit inaccuracies.}
\label{Failure cases}
\vspace{-2em}
\end{figure*}

\section*{C. Limitations and Future Works}
\textbf{Limitations:} We observe that failures mainly occur in extreme occlusion scenarios. Failure cases in extreme-occlusion interaction scenarios are shown in Fig. \ref{Failure cases}. In cases where both the hand and object are severely occluded, the model struggles to capture sufficient visual cues for accurate spatial reasoning, often resulting in erroneous pose estimations. Particularly, when the object is entirely obscured by the hand, the lack of observable features significantly hampers the model ability to infer object orientation and position. These limitations highlight the need for future work on enhancing the model ability to reason under severe occlusions.

\noindent \textbf{Future Work:}
To address the limitation under severe occlusion, future work will focus on incorporating temporal cues from previous interaction frames together with physical and motion constraints to better infer occluded hand-object interaction relationship. In addition, we aim to extend the evaluation to both third-person and egocentric views by constructing a multi-view dataset that integrates third-person data and egocentric recordings. Training on multi-view data will enhance the model viewpoint adaptability, and we will further explore multi-view feature fusion to improve hand-object interaction pose estimation accuracy.


%
%

\end{document}